\documentclass[11pt]{article}

\usepackage{authblk}
\usepackage{fullpage}

\usepackage[usenames,dvipsnames]{xcolor}
\usepackage{xspace}
\usepackage{researchpack}
\usepackage{tikz}
\usetikzlibrary{decorations.pathreplacing}
\usetikzlibrary{arrows}
\usetikzlibrary{patterns}

\newcommand{\dist}{\ensuremath{\mathrm{dist}}\xspace}
\newcommand{\reg}{\ensuremath{\mathrm{reg}}\xspace}

\newcommand{\method}{\textsc{missle}\xspace}
\newcommand{\METHOD}{MISSLE\xspace}

\newcommand{\defeq}{\ensuremath{:=}\xspace}
\newcommand{\program}{\ensuremath{M}\xspace} 
\newcommand{\dataset}{\ensuremath{S}\xspace}

\newcommand{\context}{\ensuremath{\psi}\xspace}

\newcommand{\neighbour}{\ensuremath{\calN}\xspace}

\newcommand{\Sol}{\ensuremath{\calX^*}\xspace}

\newtheorem{definition}{Definition}

\newtheorem{example}{Example}

\title{Learning Mixed-Integer Linear Programs from Contextual Examples}
\author{Mohit Kumar$^{1}$}
\author{Samuel Kolb$^{1}$}
\author{Luc De Raedt$^{1}$}
\author{Stefano Teso$^{2}$}
\affil{
    {$^{1}$KU Leuven, Belgium \hfill \\ \texttt{name.surname@cs.kuleuven.be}} \\
    {$^{2}$University of Trento, Italy \hfill \\ \texttt{stefano.teso@unitn.it}}
}

\begin{document}
\maketitle

\begin{abstract}
Mixed-integer linear programs (MILPs) are widely used in artificial intelligence and operations research to model complex decision problems like scheduling and routing.
Designing such programs however requires both domain and modelling expertise.
In this paper, we study the problem of acquiring MILPs from \emph{contextual} examples, a novel and realistic setting in which examples capture solutions and non-solutions within a specific context.
The resulting learning problem involves acquiring continuous parameters -- namely, a cost vector and a feasibility polytope -- but has a distinctly combinatorial flavor.
To solve this complex problem, we also contribute \method, an algorithm
for learning MILPs from contextual examples.
\method uses a variant of stochastic local search that is guided by the gradient of a continuous surrogate loss function.
Our empirical evaluation on synthetic data shows that \method acquires better MILPs faster than alternatives based on stochastic local search and gradient descent.
\end{abstract}

\section{Introduction}
\label{intro}

Mixed-integer linear programs (MILPs) are widely used in operations research to model decision problems like personnel rostering~\cite{nsp}, timetabling~\cite{tt}, and routing~\cite{tsp}, among many others.
%
A MILP is a constrained optimization program of the form:
\begin{align}
    \textstyle
    \max_{\vx}
        & \quad \vc^\top \vx \label{eq:cx}
    \\
    \mathrm{s.t.}
        & \quad \vA \vx \le \vb \label{eq:Axb}
\end{align}
over variables $\vx = (x_1, \ldots, x_n)$.  Like in regular linear programs, the objective function is linear with parameter $\vc \in \bbR^n$ (Eq.~\ref{eq:cx}) and the feasible set is a polytope defined by a coefficient matrix $\vA \in \bbR^{m \times n}$ and a bias vector $\vb \in \bbR^m$ (Eq.~\ref{eq:Axb}).
What makes MILPs special is that some of the variables $\vx$ are restricted to be integers.  This allows MILPs to capture numerical optimization problems with a strong combinatorial component.  For instance, $0$--$1$ variables can be used to decide whether an action should be taken or not, like taking a particular route, buying a machine, or assigning a nurse to a shift.
While MILPs are NP-hard to solve in general, practical solvers like Gurobi~\cite{gurobi} and CPLEX~\cite{cplex} can readily handle large instances, making MILPs the formalism of choice in many scientific and industrial  applications.

Designing MILPs, however, requires substantial modelling expertise, making it hard for non-experts to take advantage of this powerful framework.
One option is then to use constraint learning~\cite{de2018learning} to induce MILPs from historical data, e.g., examples of high-quality solutions (positive examples) and sub-par or infeasible configurations (negative examples).
A major difficulty is that in practice historical data is collected under the effects of temporary restrictions and resource limitations.  For instance, in a scheduling problem some workers may be temporarily unavailable due to sickness or parental leave.  We use the term ``context'' to indicate such temporary constraints.  Crucially, by restricting the feasible space, contexts can substantially alter the set of optimal solutions, biasing the data, cf. Figure~\ref{fig:example}.  Learning algorithms that ignore the contextual nature of their supervision are bound to perform poorly~\cite{kumar2020learning}.  Existing approaches for learning MILP programs suffer from this issue.

The aim of this work is to amend this situation.  In particular, we formalize contextual learning of MILPs and show that the resulting learning problem contains both continuous and combinatorial elements that make it challenging for standard techniques.  To solve this issue, we introduce \method, a novel approach that combines ideas from combinatorial search and gradient-based optimization.  In particular, \method relaxes the original loss to obtain a natural, smoother surrogate on which gradient information can be readily computed, and then uses the latter to guide a stochastic local search procedure.  Our empirical evaluation on synthetic data shows that \method performs better than two natural competitors -- namely, pure gradient descent and pure stochastic local search -- in terms of quality of the acquired MILPs and computational efficiency.

\section{Learning MILPs from Contextual Data}

\begin{figure*}[t]
    \centering

    \begin{tikzpicture}
    
    \begin{scope}[xshift=0,scale=1.5]

            \fill[opacity=0.1] (0.5, 0.1) -- (1.7, 0.1) -- (1.7, 0.8) -- (0.5, 1.0) -- (0.5, 0.1);
            \draw[thick,gray] (0.5, 0.1) -- (1.7, 0.1) -- (1.7, 0.8) -- (0.5, 1.0) -- (0.5, 0.1);

            \foreach \x in {0.2, 0.4, ..., 1.8} {
                \foreach \y in {0.2, 0.4, ..., 1} {
                    \fill[color=black] (\x, \y) circle (0.01);
                }
            }

            \draw[thick,color=black,->] (0.0, 0.0) -- (0.13, 0.13) node[left]{$\;\;$opt. dir.} -- (0.18, 0.18);

            \fill[color=SeaGreen] (1.6, 0.8) circle (0.04);

            \draw[->] (0, 0) -- (1.8, 0) node[right] {$x_1$};

            \draw[->] (0, 0) -- (0, 1) node[left] {$x_2$};

        \end{scope}

        \begin{scope}[xshift=11em,scale=1.5]

            \fill[opacity=0.1] (0.5, 0.1) -- (1.7, 0.1) -- (1.7, 0.8) -- (0.5, 1.0) -- (0.5, 0.1);
            \draw[thick,gray] (0.5, 0.1) -- (1.7, 0.1) -- (1.7, 0.8) -- (0.5, 1.0) -- (0.5, 0.1);

            \foreach \x in {0.2, 0.4, ..., 1.8} {
                \foreach \y in {0.2, 0.4, ..., 1} {
                    \fill[color=black] (\x, \y) circle (0.01);
                }
            }

            \draw[very thick,NavyBlue] (-0.1, 0.55) -- (1.9, 0.55) node[right] {$\psi$};
            \draw[very thick,NavyBlue,->] (0.3, 0.55) -- (0.3, 0.35);
            \fill[pattern=north east lines, pattern color=NavyBlue,opacity=0.4] (0.5,0.1) rectangle (1.7,0.55); 

            \fill[color=Red] (1.6, 0.4) circle (0.04);

            \draw[->] (0, 0) -- (1.8, 0) node[right] {$x_1$};

            \draw[->] (0, 0) -- (0, 1) node[left] {$x_2$};

        \end{scope}
        
        \begin{scope}[xshift=22em,scale=1.5]

            \fill[opacity=0.1] (0.5, 0.1) -- (1.7, 0.1) -- (1.7, 0.8) -- (0.5, 1.0) -- (0.5, 0.1);
            \draw[thick,gray] (0.5, 0.1) -- (1.7, 0.1) -- (1.7, 0.8) -- (0.5, 1.0) -- (0.5, 0.1);

            \foreach \x in {0.2, 0.4, ..., 1.8} {
                \foreach \y in {0.2, 0.4, ..., 1} {
                    \fill[color=black] (\x, \y) circle (0.01);
                }
            }

            \draw[very thick,NavyBlue] (1.15, -0.1) -- (1.15, 1.1) node[right] {$\psi$};
            \draw[very thick,NavyBlue,->] (1.15, 0.3) -- (0.95, 0.3);
            \fill[pattern=north east lines, pattern color=NavyBlue,opacity=0.4] (0.5,0.1) -- (1.15,0.1) -- (1.15,0.9) -- (0.5,1.0); 

            \fill[color=Red] (1.0, 0.8) circle (0.04);

            \draw[->] (0, 0) -- (1.8, 0) node[right] {$x_1$};

            \draw[->] (0, 0) -- (0, 1) node[left] {$x_2$};

        \end{scope}

    \end{tikzpicture}
    
    \caption{Toy MILP and two contexts.  Left:  MILP over two integer variables $(x_1, x_2)$.  The feasible polytope is in gray, the optimization direction goes top-right.  The optimum is in green.  Middle: Once a context $\context = x_2 \le t_2$ (blue) is added, the optimum changes substantially (red).  Right: same for $\context' = x_1 \le t_1$.  A MILP learned using the three optima as positive examples while ignoring their contextual nature would be forced to consider all points $(x_1, x_2)$ in the plane as having the same cost (i.e., $c_1 = c_2$), which is clearly undesirable.}
    \label{fig:example}
\end{figure*}
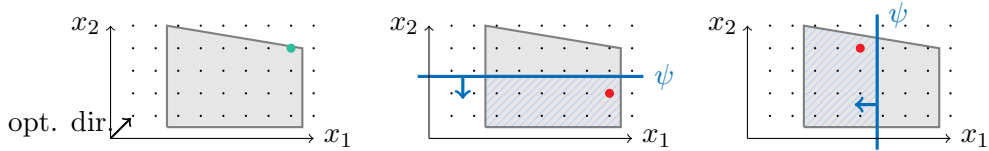

Our goal is to acquire a MILP -- and specifically its parameters $\vA$, $\vb$, and $\vc$ -- from examples of contextual solutions and non-solutions.  Contexts can alter the set of solutions of a MILP:  a configuration $\vx$ that is optimal in context $\context$ might be arbitrarily sub-optimal or even infeasible in a different context $\context'$, cf.~Figure~\ref{fig:example}.  To see this, consider the following example:

\begin{example}
\label{ex:manufacturing}
A company makes two products ($P_1$ and $P_2$) using two machines ($M_1$ and $M_2$). Producing one unit of $P_1$ requires $40$ minutes of processing time on machine $M_1$ and $20$ minutes on machine $M_2$, while each unit of $P_2$ takes $30$ minutes on $M_1$ and $40$ minutes on $M_2$.  Both machines can be run for a maximum of 4 hours every day. The company makes a profit of $\$ 20$ on each unit of $P_1$ and $\$ 18$ on each unit of $P_2$. The aim is to decide the number of units $\vx = (x_1, x_2)$ to produce for each product so that the profit is maximised.
This problem can be written as a MILP:
\begin{align*}
    \max & \quad  20x_1 + 18x_2 \\
    \mathrm{s.t.} & \quad 40x_1 + 30x_2 \le 240\\
     & \quad 20x_1 + 40x_2 \le 240
\end{align*}
The optimal solution for this problem is $\vx = (3, 4)$, giving a total profit of $\$132$. Now, consider a temporary reduction in demand for $P_2$ and hence the company wants to produce maximum $3$ units every day. This context can be added as a hard constraint: $x_2\le 3$, leading to a different optimal solution, namely $\vx = (6, 0)$. Notice that the solution without context is infeasible in this context.
\end{example}
We define the problem of learning MILPs from contextual examples as follows:
\begin{definition}
    Given a data set of context-specific examples $\dataset \defeq \{(\context_k, \vx_k, y_k) \,:\, k = 1, \ldots, s\}$, where $y_k \in \{0, 1\}$ is $1$ iff $\vx_k$ is a high-quality solution for context $\context_k$,
    find a MILP with parameters $\theta = (\vA, \vb, \vc)$ that can be used to obtain high-quality configurations in other contexts of interest.
\end{definition}
\noindent
By ``high-quality'' configurations, we mean configurations that are feasible and considered (close to) optimal.  Going forward we will use $\program(\theta)$ to indicate the MILP defined by $\theta$.

\subsection{Contexts}

In the following, we will restrict ourselves to handling contexts that impose additional \emph{linear} constraints $\vS \vx \le \vt$ to the problem.  This choice is both convenient and flexible.  It is convenient because injecting linear contexts into a MILP gives another MILP, namely:
\begin{align}
    \textstyle
    \max_{\vx}
        & \quad \vc^\top \vx \\
    \mathrm{s.t.}
        & \quad \vA \vx \le \vb, \quad \vS \vx \le \vt
\end{align}
It is also very flexible:  the polytope $\vS \vx \le \vt$ can naturally encode frequently used constraints like partial assignments (which fix the value of a subset of the variables) and bounding-box constraints (which restrict the domain of a subset of the variables to a particular range).  Furthermore, it is often possible to encode complex non-linear and logical constraints into this form using linearization techniques~\cite{vielma2015mixed}.

\subsection{An initial formulation}

Learning a MILP amounts to searching for a program $\program(\theta)$ that fits the data well.  Naturally, ignoring the contextual nature of the data -- for instance, by treating all positive examples as if they were global, rather than contextual, optima -- can dramatically reduce the quality of the learned program~\cite{kumar2020learning}, cf. Figure~\ref{fig:example}.

Given a MILP $\program(\theta)$ and a context $\context$, we define $h_\theta(\vx, \context)$ to be a binary classifier that labels configuration $\vx$ as positive if and only if it is solution to $\program(\theta)$ in context $\context$:
\[
    h_\theta(\vx,\context) \defeq \Ind{\vx \in \Sol(\theta, \context)}
\]
Here $\Ind{cond}$ is the indicator function that evaluates to $1$ if $cond$ holds and to $0$ otherwise, and $\Sol(\theta, \context)$ is the set of contextual solutions of $\program(\theta)$ in context $\context$.
Given a dataset $\dataset$ of contextual examples, we propose to learn a MILP by minimizing the following 0-1 loss:
\[
    \textstyle
    L_{\dataset}(\theta) \defeq \frac{1}{s} \sum_{k=1}^s \Ind{h(\vx_k,\context_k) \ne y_k}
    \label{eq:zero-one-loss}
\]
This amounts to looking for a MILP $\program(\theta)$ that minimizes the number of misclassified training examples. 

\subsection{Advantages and Limitations}

The above strategy is motivated by the work of Kumar~\emph{et~al.}~\cite{kumar2020learning}, who have shown that, under suitable assumptions, classifiers $h_\theta$ that make few mistakes on the training set correspond to programs $\program(\theta)$ that output high-quality configurations in (even unobserved) target contexts.\footnote{These results were obtained for MAX-SAT programs but do carry over to general constraint optimization programs.}

A downside of this formulation is that it is computationally challenging.
Solving the above optimization problem is equivalent to minimizing the $0$--$1$ loss on the training set, which is known to be NP-hard~\cite{ben2003difficulty}. 
Furthermore, the objective function is piece-wise constant:  there exist infinitely many pairs of MILP models $\theta^1 \ne \theta^2$ that have the same set of solutions and that therefore lie on the same plateau of the empirical loss.  This makes it hard to apply optimization procedures based on pure gradient descent, as the gradient is constant almost everywhere.  Standard local search is also not very effective, as it proceeds by making small uninformed steps and therefore might spend a long time trapped on a plateau.
A second issue is that evaluating the loss is hard:  checking whether an instance $\vx_k$ is predicted as positive -- that is, evaluating $h(\vx_k, \context_k) = 1$ -- involves finding an optimum $\vx^* \in \Sol(\theta; \context)$ and ensuring that it has the same value as $\vx_k$.  The issue is that computing $\vx^*$ requires to solve the candidate MILP in context $\context_k$, which is NP-hard in general.  This step needs to be carried out repeatedly when minimizing Eq.~\ref{eq:zero-one-loss}, so it is important to keep the number of optimization steps at a minimum.

\section{Learning MILPs with \METHOD}
\label{sec:method}

Due to the difficulty of optimizing Eq.~\ref{eq:zero-one-loss} directly, we employ a smoother surrogate loss that offers gradient information and use it to guide a stochastic local search procedure.

\subsection{A Surrogate Loss}

We build a surrogate loss by enumerating the various ways in which $\program(\theta)$ can \emph{mislabel} a contextual example and defining a loss for each case.  Below, we write $\dist_j(\vx)$ to indicate the Euclidean distance between $\vx$ and the $j$th hyperplane $(\va_j, b_j)$ of the feasible polytope~\cite{scholkopf2002learning}:
\[
    \dist_j(\vx) \defeq |\va_j^\top \vx - b_j|\ / \norm{\va_j}_2
\]
Additionally, we write $\reg_{\theta}(\vx_k)$ to indicate the regret of $\vx_k$ in context $\context_k$, which measures the difference in quality between $\vx_k$ and a truly optimal configuration $\vx_k^* \in \Sol(\theta, \context_k)$ according to the \emph{learned} cost vector $\vc$:
\[
    \reg_{\theta}(\vx_k) \defeq \vc^\top\vx_k^* - \vc^\top\vx_k = \vc^\top (\vx_k^* - \vx_k)
    \label{eq:regret}
\]
%
In the remainder of this section, let $(\context_k, \vx_k, y_k)$ be a \emph{misclassified} example.  If it is positive (that is, $y_k = 1$), then it is predicted as either infeasible or suboptimal by $\program(\theta)$.
%
If it is feasible, then it is sufficient to make it optimal.  This can be achieved by either increasing the estimated cost of $\vx_k$, by excluding the actual optimum $\vx_k^*$ from the feasible region, or by bringing $\vx_k$ closer to the boundary.
This leads to the following loss:
\[
    \textstyle
    \reg_\theta(\vx_k) + \min_j \dist_j(\vx_k^*) + \min_j \dist_j(\vx_k)
    \label{eq:pos_feasible}
\]
where the $\min_j$ operations select the closest hyperplane.

On the other hand, if $\vx_k$ is positive but infeasible, then we have to enlarge the feasible region to make it feasible.  This is accomplished by penalizing the distance between $\vx_k$ and all the hyperplanes that exclude $\vx_k$ from the feasible region using the following loss:
\[
    \textstyle
    \sum_j \Ind{ \va_j^\top \vx_k > b_j } \dist_j(\vx_k)
\]

Finally, if $\vx_k$ is negative ($y_k = 0$) -- but classified as positive -- then we want to make it either sub-optimal or infeasible.  This can be achieved by increasing its cost or by adapting the feasible region so that it excludes this example, giving:
\[
    \textstyle
    \exp(-|\reg_\theta(\vx_k)|)+ \min_j \dist_j(\vx_k)
    \label{eq:neg}
\]
Note that we use exponential just to make sure that the loss is lower bounded by 0.
The full surrogate loss $\tilde{L}_S(\theta)$ is obtained by combining the various cases, as follows:
\begin{align*}
    \tilde{L}_S(\theta) \defeq \frac{1}{s} \sum_{k=1}^s &
    \begin{cases}
        \reg_\theta(\vx_k) + \min_j \dist_j(\vx_k^*) + \min_j \dist_j(\vx_k)
            \quad & \text{if}\ \vA \vx_k \le \vb \land y_k=1
        \\
        \sum_j \Ind{\va_j^\top \vx_k > b_j} \dist_j(\vx_k)
            \qquad\qquad & \text{if}\ \vA \vx_k \not\le \vb  \land y_k=1
        \\
        \exp(-|\reg_\theta(\vx_k)|)+ \min_j \dist_j(\vx_k)
            \quad & \text{if}\ \vc^\top\vx_k^* = \vc^\top \vx_k \; \land \\
            & \quad \vA \vx_k \le \vb \land y_k=0
    \end{cases}
\end{align*}
This surrogate loss is inspired by but not identical to the one given by Paulus~\emph{et~al.}~\cite{paulus2020fit}.  There are three core differences.  First, Eq.~\ref{eq:pos_feasible} includes a term $\min_j \dist_j(\vx_k)$ whose aim is to move the boundaries closer to $\vx_k$, as an optimal solution for MILP lies close to the boundary.  Second, our surrogate loss includes a term for mis-classified negative examples, integrated through Eq.~\ref{eq:neg}. Finally, we integrate the contextual information in the surrogate loss by defining $\vx_k^*$ as an optimal solution in the context $\context_k$. Apart from these technical differences, another major difference is in the setting in which we use this loss function. Paulus~\emph{et~al.} focus on end-to-end learning, for instance learning constraints from given textual descriptions. However, our focus is to use this loss to learn MILPs from contextual examples.

\subsection{Searching for a MILP}

The goal of \method is to find a program $\program(\theta)$ that minimizes the 0-1 loss defined in Eq~\ref{eq:zero-one-loss}. However, the major barrier here is the continuous search space. To tackle this we propose a variant of SLS guided by the gradient of the surrogate loss defined above.

The backbone of all SLS strategies is a heuristic search procedure that iteratively picks a promising candidate in the neighborhood $\neighbour(\theta)$ of the current candidate $\theta$, while injecting randomness in the search to escape local optima.
The challenge is how to define a neighborhood that contains promising candidates and allows the search algorithm to quickly reach the low-loss regions of the search space.  To this end, we exploit the fact that the surrogate loss $\tilde{L}_S$ is differentiable w.r.t. $\theta$ to design a small set of promising neighbors, as follows.
Each $\theta$ has three neighbours defined by the following moves:
\begin{itemize}

    \item Perturb the cost vector $\vc$ by taking a single gradient descent step of length $\lambda$ while leaving $\vA$ and $\vb$ untouched:
    $
        \vc \gets \vc - \lambda \nabla_{\vc} \tilde{L}_S(\theta)
    $.

    \item Similarly, rotating the hard constraints by performing gradient descent w.r.t. $\vA$:
    $
        \vA \gets \vA - \lambda \nabla_{\vA} \tilde{L}_S(\theta)
    $.

    \item Translate the hard constraints by updating the offsets $\vb$:
    $
        \vb \gets \vb - \lambda \nabla_{\vb} \tilde{L}_S(\theta)
    $.

\end{itemize}
In each iteration of the SLS procedure, the next hypothesis is chosen greedily, picking the candidate $\theta' \in \neighbour(\theta, \dataset, \lambda)$ with minimal \emph{true} loss $L_{\dataset}(\theta')$. Notice that this is different from the gradient step, which is based on the surrogate loss $\tilde{L}_S$ instead.
The intuition is twofold.  On the one hand, the gradient is cheap to compute and points towards higher-quality candidates, while on the other the SLS procedure smartly picks among the candidates bypassing any approximations induced by the surrogate loss.
Another advantage of this solution is that it naturally supports exploring continuous search spaces, whereas most SLS strategies focus on discrete ones.

We make use of two other minor refinements.  First, the learning rate $\lambda$ is adapted during the search, and more specifically it is set close to 1 when the loss is high and slowly decreased towards 0 as the loss decreases, more formally:
$
\textstyle
\lambda = \frac{1}{2}\ / \sqrt{\max\{100 \cdot (1-L_S(\theta)), 1\}}
$.
The idea is to take larger steps so to quickly escape high-loss regions while carefully exploring the more promising areas.
We also normalize both the direction of optimisation $\vc$ and the hyperplanes $\vA$ and $\vb$ after each step, thus regularizing the learning problem.

\subsection{\METHOD}

The pseudo-code for \method is given in Algorithm~\ref{alg:missle}. It starts with an initial candidate which is generated by first finding the convex hull of the positive points in the data, and then randomly picking $m$ sides of this convex hull to define the initial feasible region, while a random vector is chosen to be the direction of optimisation.
The rest of the code simply iterates through the neighbourhood picking the best candidate in each iteration, while randomly restarting with probability $p$, which helps the algorithm in avoiding a local minima.

\begin{algorithm}[h]
    \caption{\label{alg:missle} The \method algorithm.  Inputs: data set $\dataset$, max constraints $m$, learning rate $\lambda$, restart prob. $p$, cutoff $t$, max accuracy $\alpha$.}
    \begin{algorithmic}
        \State $\theta \gets initialCandidate(\dataset, m)$, \; $\theta_\text{best} \gets \theta$
        \While {$\mathrm{runtime} < t$ \text{and} $\mathrm{accuracy}(\theta, \dataset) < \alpha$}
            \State \textbf{with probability $p$ do}
                \State \qquad $\theta \gets initialCandidate(\dataset, m)$ \Comment{restart}
            \State \textbf{otherwise}
                \State \qquad $\theta \gets \argmin_{\theta' \in \neighbour(\theta, \dataset, \lambda)} L_{\dataset}(\theta')$ \Comment{best neighbour}
            \If{$L_{\dataset}(\theta) < L_{\dataset}(\theta_\text{best})$}
                \State $\theta_\text{best} \gets \theta$ \Comment{track best-so-far}
            \EndIf
        \EndWhile
        \State \Return $\theta_\text{best}$
    \end{algorithmic}
\end{algorithm}

\section{Experiments}

We answer empirically the following research questions:
\begin{itemize}

    \item \textbf{Q1}: How does the performance change over time?
    
    \item \textbf{Q2}: Does \method learn high quality models?
    
    \item \textbf{Q3}: How does \method compare to pure SGD and SLS approaches?

\end{itemize}
To answer these questions, we used \method to recover a set of synthetic MILPs from contextual examples sampled from them and compared its hybrid search strategy to two natural baselines:  stochastic local search (SLS) and stochastic gradient descent (SGD). The baseline using SGD can be seen as an extension of the ideas presented by Paulus~\emph{et~al.} to learn in our setting.

\subsection{Experiment Description}

To generate the training data, we first randomly generated 5 different ground-truth models $\program^*$, each with 5 variables and 5 hard constraints. For each $\program^*$, a dataset \dataset was collected by first sampling 250 random contexts and then taking 1 positive and 2 negative examples from each context. For each model $\program^*$, we generated 5 different set of data by changing the seed of randomization.

The quality of the learned model $\program$ was captured by measuring recall, precision, infeasibility, and regret.  Recall tells us what percentage of the true feasible region is covered by the learned model, while precision tells us what part of the learned feasible region is actually correct. Infeasibility gives us the percentage of the optimal solutions in $\program$ that are infeasible in $\program^*$, while regret measures the quality of the optima in $\program^*$ as defined in Equation~\ref{eq:regret}. Naturally, a better model is characterised by higher value of recall and precision and lower value of infeasibility and regret.

\begin{figure*}[tb]
    \centering
        \includegraphics[width=\linewidth]{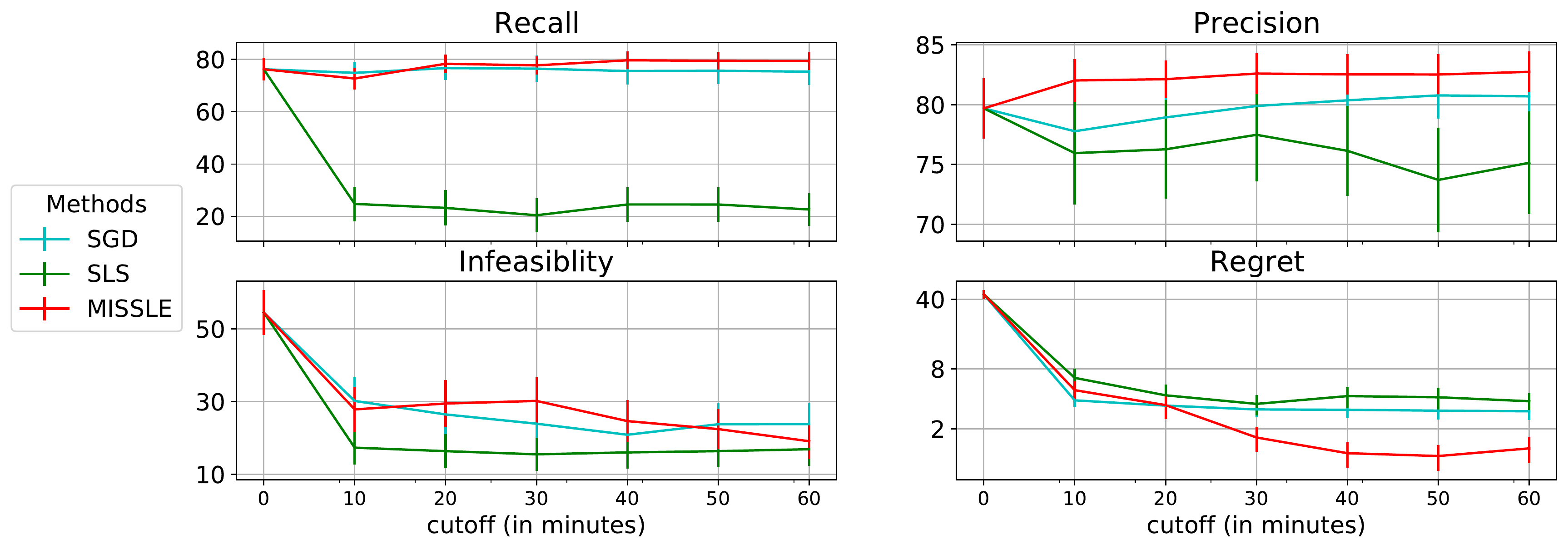}
        \caption{Combining gradient and SLS in \method leads to better performance (higher recall and precision, lower regret) compared to using them individually. The only exception is that SLS shows better performance in terms of infeasibility, however, this is because it learns a very restrictive model, as is suggested by the low recall value.}
    \label{fig:results}
\end{figure*}

\subsection{Results and Discussion}

The results are shown in Figure~\ref{fig:results}. The x-axis in each plot represents the cutoff time, which is the maximum time an algorithm is allowed to run before producing a learnt model. The y-axis for regret is logarithmic to make the differences more clear. As expected, increasing the cutoff time leads to better results, specially in terms of infeasibility and regret. With a cutoff time of 60 minutes, \method achieves $80\%$ recall and $83\%$ precision with infeasibility $\sim 20\%$ and regret $\sim 2\%$, hence $Q2$ can be answered in affirmative. To answer $Q3$, we compared \method with the two baselines based on SGD and SLS, the comparison clearly shows that \method outperforms both baselines. The only exception is that SLS shows better performance in terms of infeasibility when the cutoff time is low, however, this is because it learns a very restrictive model, as is suggested by the low recall value. SLS also has more than three times the regret attained by \method. Hence, it can be clearly concluded that combining SLS and SGD in \method leads to better performance compared to using each individually.

\section{Related Work}
\label{sec:rw}

\newcommand{\etal}{\emph{et~al.}\xspace}

Our work is motivated by advancements in constraint learning~\cite{de2018learning} and in particular by Kumar~\etal~\cite{kumar2020learning}, who introduced the problem of learning MAX-SAT models from contextual examples.  Our work extends these ideas to continuous-discrete optimization.  Kumar~\etal cast the learning problem itself as a MILP.  This strategy, while principled, can be computationally challenging.  In contrast, \method leverages a hybrid strategy that combines elements of combinatorial and gradient-based optimization.  One advantage is that \method naturally supports anytime learning.

Typical constraint learning methods~\cite{de2018learning} -- like ConAcq~\cite{bessiere2017constraint} and ModelSeeker~\cite{beldiceanu2012model} -- focus on acquiring constraints in higher-level languages than MILP, but lack support for cost functions.  Those approaches that do support learning soft constraints require the candidate constraints to be enumerated upfront, which is not possible in our discrete-continuous setting~\cite{rossi2004acquiring}.

Most approaches to learning MILPs from data either learn the cost function or the hard constraints, but not both.
Pawlak and Krawiec~\cite{pawlak2017automatic} acquire the feasibility polytope from positive and negative examples by encoding the learning problem itself as a MILP.
Schede~\emph{et~al.}~\cite{schede2019learning} acquire the feasibility polytope using a similar strategy, but -- building on work on learning Satisfiability Modulo Theory formulas~\cite{kolb2018learning} and decision trees~\cite{bertsimas2017optimal} and on syntax-guided synthesis~\cite{alur2018search} -- implement an incremental learning loop that achieves much improved efficiency.
Approaches for learning polyhedral classifiers find a convex polytope that separates positive and negative examples, often by a large margin~\cite{astorino2002polyhedral,manwani2010learning,kantchelian2014large,gottlieb2018learning}, but do not learn a cost function.  This is not a detail:  negative examples in their classification setting are known to be infeasible, while in ours they can be either infeasible or suboptimal.  This introduces a credit attribution problem that these approaches are not designed to handle.

There are two notable exceptions.  One is the work of Tan~\etal~\cite{tan2020learning}, which acquires linear programs using stochastic gradient descent (SGD).  Their technique, however, requires to differentiate through the solver, and this cannot be done for MILPs.
A second one is the work of Paulus~\emph{et~al.}~\cite{paulus2020fit}, which learns integer programs from examples using SGD and a surrogate loss.  \method uses a similar surrogate loss that however explicitly supports negative examples, as discussed in Section~\ref{sec:method}.  Another major difference is that -- as shown by our experiments -- our hybrid optimization strategy outperforms pure SGD.

Most importantly, none of the works mentioned above support contextual examples.  Because of this, they are bound to learn sub-par MILPs when applied to contextual data.

\section{Conclusion}

We introduced the problem of learning MILPs from contextual examples as well as \method, a practical approach that combines stochastic local search with gradient-based guidance.  Our preliminary evaluation shows that \method outperforms two natural competitors in terms of scalability and model quality.

This work can be improved and extended in several directions.  First and foremost, a more extensive evaluation is needed.
Second, \method should be sped up by integrating the incremental learning strategy of~\cite{schede2019learning,kolb2018learning} in which a model is learned from progressively larger subsets of examples, stopping whenever the improvement in score saturates.
Third, it will be interesting to see the attribution of positive and negative examples in the learning process, we believe positive examples carry much more information compared to a negative example, hence it might be beneficiary to learn from only positive examples, however this needs more extensive experiments.
Finally, if meta-information about which negatives are infeasible and which ones are sub-optimal is available, however this information is often not known and thus \method does not use this information, we will work on a new algorithm that can utilise this extra information and check the impact on the performance.

\subsection*{Acknowledgments}

This work has received funding from the European Research Council (ERC) under the European Union’s Horizon 2020 research and innovation programme (grant agreement No.  [694980] SYNTH: Synthesising Inductive Data Models).  The research of Stefano Teso was partially supported by TAILOR, a project funded by EU Horizon 2020 research and innovation programme under GA No 952215.

\bibliography{paper}
\end{document}